\documentclass{article}

\usepackage[utf8]{inputenc} 
\usepackage[T1]{fontenc}    
\usepackage{hyperref}       
\usepackage{url}            
\usepackage{booktabs}       
\usepackage{amsfonts}       
\usepackage{nicefrac}       
\usepackage{microtype}      
\usepackage{lipsum}
\usepackage{graphicx}
\graphicspath{ {./images/} }
\usepackage{float}
\usepackage{lmodern} 
\usepackage{amsmath}
\title{Computer Vision-Based Vehicle Allotment
System using Perspective Mapping}
\author{
\makebox[\textwidth][c]{
\begin{tabular}{c c}
\textbf{Sonakshi Satapathy$^{1}$} & \textbf{Prachi Nandi$^{1}$} \\
National Institute of Technology Rourkela & National Institute of Technology Rourkela \\
\texttt{sonakshi1901@gmail.com} & \texttt{prachinandi237@gmail.com} \\
\\
\multicolumn{2}{c}{\textbf{Suchismita Chinara$^{2}$}} \\
\multicolumn{2}{c}{National Institute of Technology Rourkela} \\
\multicolumn{2}{c}{\texttt{suchismita@nitrkl.ac.in}}
\end{tabular}
}
}
\begin{document}
\maketitle

\begin{abstract}
Smart city research envisions a future in which data-driven
solutions and sustainable infrastructure work together to define urban
living at the crossroads of urbanization and technology. Within this
framework, smart parking systems play an important role in reducing
urban congestion and supporting sustainable transportation. Automating parking solutions have considerable benefits, such as increased efficiency and less reliance on human involvement, but obstacles such as
sensor limitations and integration complications remain. To overcome
them, a more sophisticated car allotment system is required, particularly in heavily populated urban areas. Computer vision, with its higher
accuracy and adaptability, outperforms traditional sensor-based systems
for recognizing vehicles and vacant parking spaces. Unlike fixed sensor
technologies, computer vision can dynamically assess a wide range of visual inputs while adjusting to changing parking layouts. This research
presents a cost-effective, easy-to-implement smart parking system utilizing computer vision and object detection models like YOLOv8. Using
inverse perspective mapping (IPM) to merge images from four camera
views, we extract data on vacant spaces. The system simulates a 3D
parking environment, representing available spots with a 3D Cartesian
plot to guide users. 
\end{abstract}


\section{Introduction}
Smart parking systems \cite{rp1} \cite{rp36} identify the availability of parking spaces in real-time, helping to optimize traditional parking systems, They are characterized as creative solutions which utilize modern technology to maximize parking space utilization and enhance user convenience. In contrast to conventional parking systems, which frequently include inefficiencies like poor sight of available spaces, traffic jams in parking lots, and drawn-out search procedures, smart parking systems combine a number of technologies to solve these problems. These systems lead vehicles to available spots quickly, give real-time information about parking availability, and simplify the parking process overall by utilizing sensors, cameras, and data analytics. Smart parking systems greatly enhance traffic flow, lessen congestion, and promote a more sustainable urban environment by automating and digitizing parking operations using computer vision-based techniques.
Computer vision-based systems \cite{rp17} are a subset of artificial intelligence that allows machines to analyze and understand visual information from their surroundings. In the parking system scenario, a variety of computer vision techniques, such as object identification, picture segmentation, and optical character recognition (OCR) \cite{rp30} \cite{rp31}, can be used. However, for this project, the focus is on object detection. Object detection algorithms enable the precise localization and identification of objects within images or video frames, making them well-suited for tasks such as identifying vacant parking spaces and monitoring vehicle movements in parking lots. We will be using YOLOv8 (You Only Look Once) for detecting vehicles and pillars. \\
YOLO (You Only Look Once) \cite{rp15} is a pioneering object detection technique that transformed computer vision applications by enabling real-time detection. Unlike typical detection systems, which require numerous passes over an image, YOLO processes the entire image in one step, making it extremely fast and efficient. The latest generation, YOLOv8, expands on this foundation by improving accuracy, speed, and variety. It includes sophisticated features like as anchor boxes for better localization, multi-scale prediction for managing objects of varying sizes, and a stronger backbone network for improved recognition. These improvements make YOLOv8 an excellent candidate for object identification tasks, especially in dynamic contexts such as smart parking systems, where real-time accuracy and speed are critical. After object detection, we need to extract the empty parking spaces through inverse mapping. 
Inverse mapping \cite{rp20}, also known as inverse perspective mapping (IPM), is an important approach in computer vision for calculating the depth or distance of objects from a camera. It's especially useful in applications such as smart parking systems, where knowing the spatial layout of things is critical. By using inverse mapping, we can convert the 2D coordinates of objects acquired by the camera into 3D coordinates, allowing us to properly establish their spatial placements. This information is important for charting the parking lot and locating open parking spaces between automobiles or obstructions. In essence, inverse mapping bridges the gap between 2D image data and 3D spatial comprehension, making it a crucial tool in smart parking systems. \\

Our inspiration comes from the difficulties encountered by indoor parking systems, specifically the high expense of conventional sensors such as PIR and ultrasonic in large, multi-story parking areas \cite{rp33} \cite{rp34} \cite{rp35}. Our goal is to create an affordable solution that gets over the budgetary obstacles that come with sensor-based systems by using CCTV video instead. Beyond cost effectiveness, we also prioritize better navigation and less traffic at entry and exit locations. It is imperative to improve navigation systems and optimize traffic flow in indoor parking facilities due to the restricted visibility and intricate layouts that frequently cause confusion and inefficiency. Our objective is to provide a novel solution that not only tackles the financial difficulties associated with indoor parking but also greatly enhances the whole parking experience, facilitating more efficient traffic management and user convenience.
This project aims to transform smart parking systems by enhancing their efficiency, accessibility, and user engagement through several strategic improvements. Firstly, it focuses on developing a computer vision-based intelligent parking system that efficiently detects parking spots using modern technologies adaptable to various camera perspectives and parking lot configurations. Secondly, it incorporates routing algorithms like Dijkstra's for real-time navigation \cite{rp29}, guiding drivers to the nearest open parking spaces within the parking complex. Thirdly, it utilizes 3D simulations to evaluate system efficacy in identifying empty spaces, ensuring accuracy and efficiency. Overall, the project aims to provide a cost-effective and practical solution for indoor parking management, enhancing mobility and convenience while prioritizing user satisfaction and optimizing the parking experience.

\section{Related Works}
Smart Parking systems use sensors, cameras, and communication technologies to detect available parking spaces and guide drivers, improving efficiency and convenience. Indoor Smart Parking systems are especially needed to provide accurate, real-time information in environments where GPS is ineffective, reducing search time and optimizing space usage. Qiang et al. \cite{rp22} proposed a visible light communication technology with LED light sources, infrared sensors, Zigbee, and Bluetooth for indoor parking space detection and navigation. On the other hand, Kianpisheh et al. \cite{rp1} introduced the Smart Parking System (SPS) with ultrasonic sensors to identify parking spaces and incorrect parking. SPS offers real-time empty space detection, incorrect parking identification, and parking availability display, enhancing user convenience. However, SPS's reliance on expensive ultrasonic sensors and potential limitations in uncovered areas can pose installation and coverage challenges, while technical issues may affect its accuracy and performance. 
Indoor smart parking systems, often integrated with 3D mapping technology, address the issues of precise navigation and real-time visualization. Bojja et al. \cite{rp23} proposed a novel method for three-dimensional navigation and localization of a land vehicle in a multistorey parking garage using low-cost gyro and odometer sensors combined with a 3D map, particle filtering, and collision detection techniques, eliminating the need for an altimeter or additional infrastructure devices. 
3D mapping in indoor parking systems can be achieved through various techniques, including LiDAR for precise distance measurement, camera-based methods for rich visual data, and deep learning for enhanced processing and feature extraction from both sensor types. Gong et al. \cite{rp24} presented a novel method for mapping and semantic modeling of underground parking lots using low-cost Backpack Laser Scanning (BLS) or LiDAR systems, consisting of a Sparse Point Cloud (SPC)-based SLAM algorithm and a modified PointNet semantic modeling algorithm, achieving centimeter-level accuracy and 84.8\% precision in semantic segmentation. For a camera-based method, Du et al. \cite{rp25} proposed a novel camera-assisted region-based magnetic field (MF) fingerprinting technique for indoor positioning, enhancing MF-based localization accuracy by integrating camera-based positioning in areas with fewer disturbances, and demonstrating superior performance compared to MF-only and camera-only systems through a system that uses image feature points for matching and displays user locations on a visual 3D map of the indoor environment. Using the deep learning-based method, Cai et al. \cite{rp26} showcased an accurate and real-time video system for Internet of Things (IoT) and smart city applications, utilizing deep convolutional neural networks (DCNNs) and a novel vehicle tracking filter to enhance accuracy by combining information across multiple video frames, achieving performance comparable to expensive sensor-based systems, and demonstrating significant scalability and rich output beyond traditional binary occupancy statistics. 
Various 3D mapping techniques provide the foundation for building rich environmental understanding. By coupling these techniques with computer vision models, objects can be detected and recognized not just by their appearance, but also by their precise 3D shape and location within the environment. Li et al. \cite{rp27} proposed a vehicle localization system for indoor parking lots, using stationary cameras with AI-based car recognition (YOLOv3) and close-range photogrammetry to determine car positions, tested in a parking lot with residents' cars. While the system performs well in terms of time and precision, localization accuracy is affected by image noise and lighting, and future improvements are needed to convert local coordinates to a geodetic coordinate system due to the lack of on-site control points and RTK devices.
There have been multiple works in the areas of computer vision models with camera-based methods for object detection in smart parking systems. Liu et al. \cite{rp20} developed a low-cost camera-based smart parking system specifically for airports, utilizing inverse perspective mapping (IPM) to generate top-down representations of parking areas and retrieve parking details. Despite its advantages in affordability and accuracy, concerns regarding privacy, and system maintenance need to be addressed for optimal performance and user satisfaction. Bibi et al. \cite{rp7}, on the other hand, proposed a Computer Vision-based method that divides a parking lot into virtual blocks and identifies automobiles within each block as regions of interest (ROI). The system uses color-coded blocks to indicate parking availability, with green for free spaces and red for reserved ones based on connected regions exceeding a threshold of eighty. Although offering an average performance of 99.5\% accuracy, the system's effectiveness may vary depending on the camera type and quality used for surveillance, potentially limiting coverage in certain areas and requiring technical expertise for implementation. While here \cite{rp28}, the authors proposed an intelligent parking guidance system that used a monocular camera, computer vision, and edge computing to detect vacant parking spaces and guide drivers to the nearest available spot via a web interface. Their system integrates real-time detection, localisation, and navigation and was later extended to a multi-camera setup.
This paper introduces an affordable, image-based approach from different camera angles to 3D indoor mapping, aiming to simplify access to 3D models of indoor parking spaces by employing deep learning methods for efficient and precise object localization within images.

\section{Proposed Work}
\subsection{Dataset Collection}
Gathering a good dataset is critical for testing and training object detection algorithms like YOLOv8, especially in complex environments like indoor parking lots. To ensure the model's accuracy and durability, these datasets must encompass a diverse range of scenarios. To address this, we propose capturing videos from a 3D simulation of the interior parking lot created with Spline.AI. This simulation depicts the parking area realistically, with differences in lighting, car kinds, parking layouts, and obstructions like pillars and walls. By collecting a variety of events, we ensure that the dataset accurately represents real-world conditions. 

\begin{figure}[htbp]
\centering
    \includegraphics[scale=0.7]{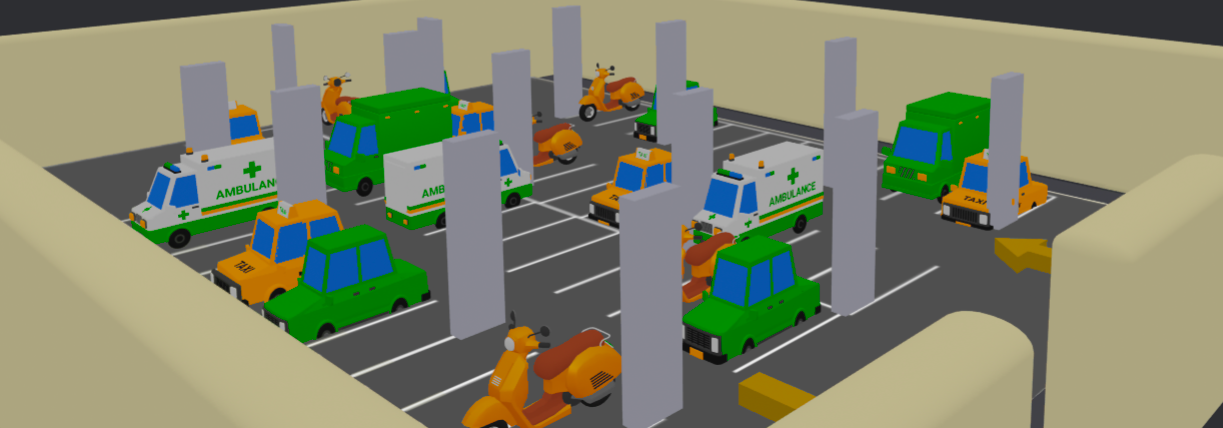}
    \caption{Indoor Parking Lot from a camera angle}
\end{figure}
\begin{figure}[htbp]
\centering
    \includegraphics[scale=0.75]{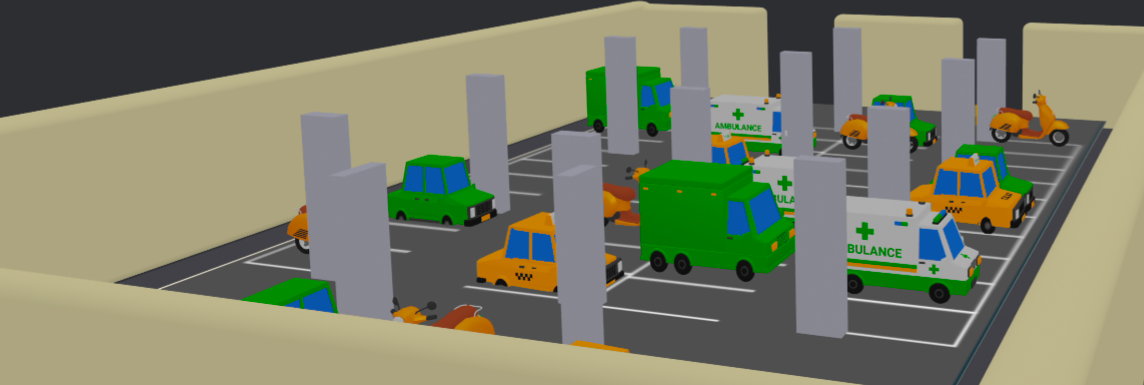}
    \caption{Indoor Parking Lot from a camera angle}
\end{figure}

\subsection{Model Architecture}
The YOLO (You Only Look Once) architecture \cite{rp32} is designed around a single convolutional neural network (CNN) that predicts bounding boxes and class probabilities directly from full images in one pass. It's divided into two main parts: the backbone network and the detection head. The backbone network typically consists of convolutional layers like Darknet or ResNet, which extract features from the input image. The detection head then takes these features and predicts bounding boxes, class probabilities, and confidence scores for each object in the image. YOLO is known for its speed and efficiency in real-time object detection tasks. \\

The YOLOv8 architecture builds upon the YOLO (You Only Look Once) framework, integrating dynamic anchor boxes that adapt to object sizes and locations within images. This dynamic anchor box mechanism significantly improves accuracy compared to previous versions like YOLOv5 and YOLOv7. YOLOv8 utilizes a series of convolutional layers for feature extraction, followed by a detection head that predicts bounding boxes, class probabilities, and confidence scores for objects. The inclusion of dynamic anchor boxes allows YOLOv8 to better handle varying object scales and positions, resulting in more precise and reliable object detection. This enhanced accuracy makes YOLOv8 a preferred choice for tasks requiring real-time and accurate object detection, in smart parking solutions.

\begin{figure}[htbp]
    \centering
    \includegraphics[scale = 0.4]{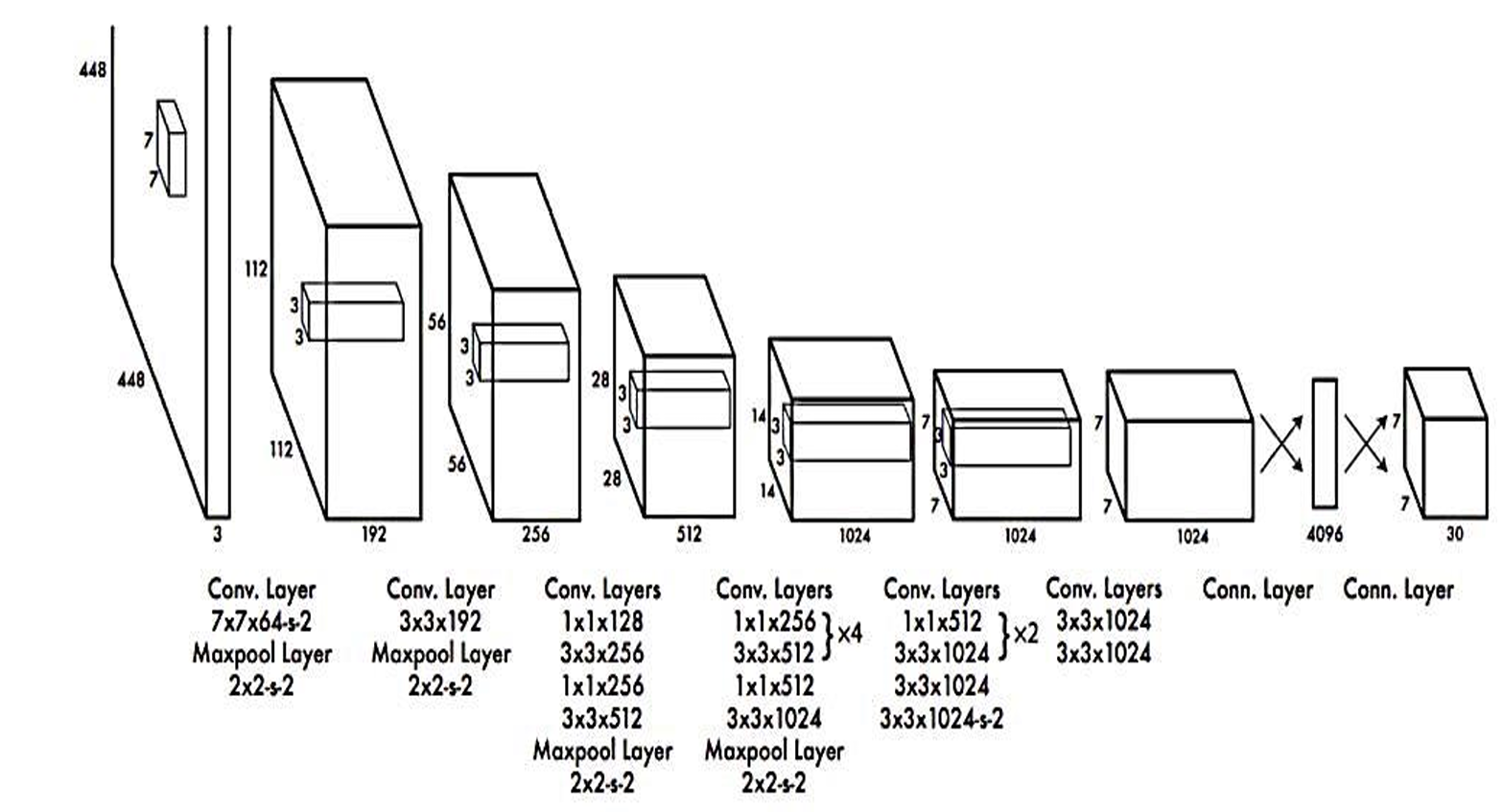}
    \caption{Model architecture of YOLO}
    \label{fig:enter-label}
\end{figure}

\subsection{System Model}
The Proposed Model process describes the entire workflow for the proposed work. It is separated into three phases, which are listed below.
\begin{enumerate}
    \item Input Stage: Provides information on data collecting and preparation.
    \item Processing Stage: Covers model pre-processing and training.
    \item Output Stage: Contains the created 3D Cartesian plot
\end{enumerate}

\begin{figure}[htbp]
    \centering
    \includegraphics[scale = 0.7]{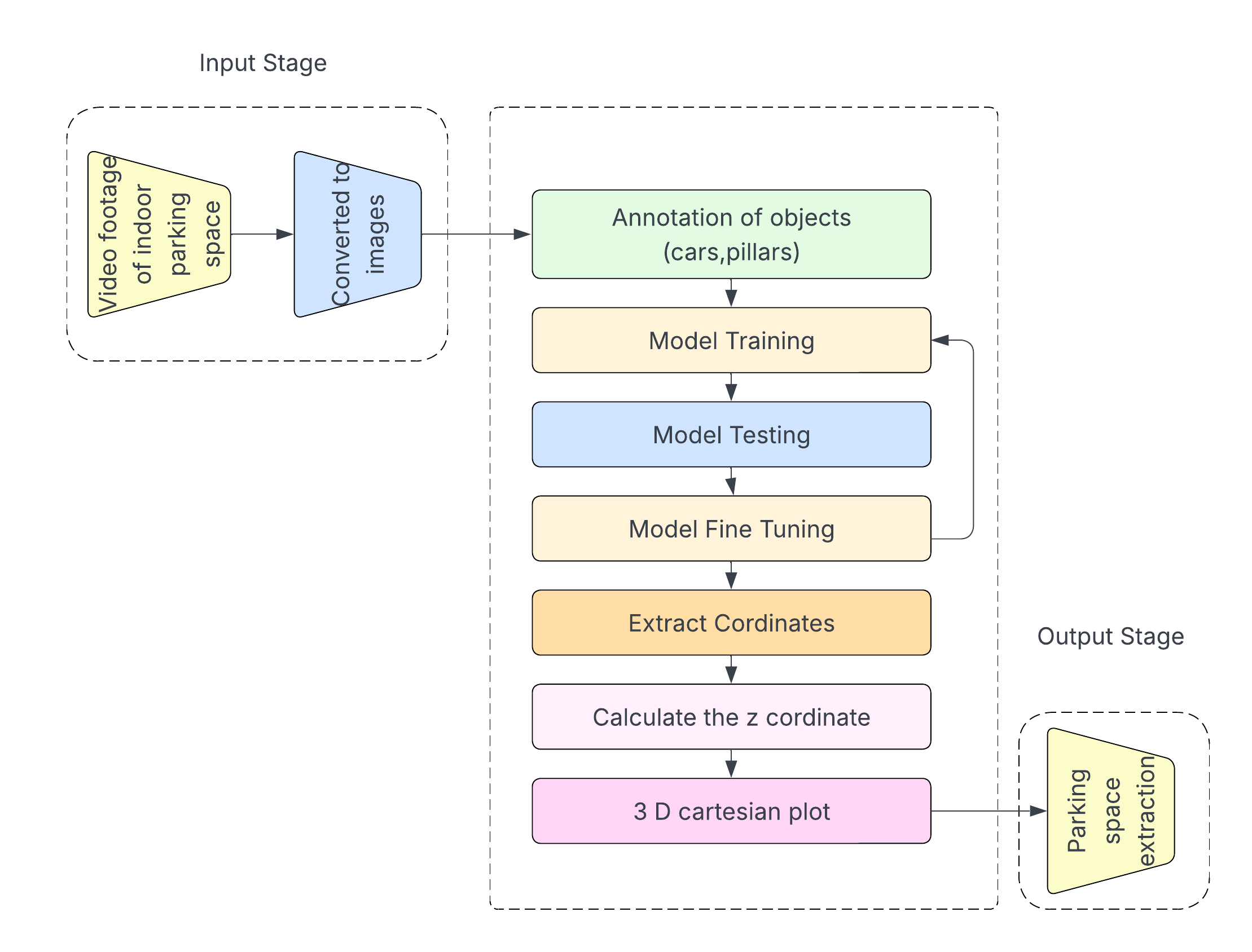}
    \caption{System Model}
    \label{fig:enter-label}
\end{figure}

\subsubsection{Input Stage:}
In our proposed methodology, we record video footage from four different camera viewpoints within a Spline.AI-generated 3D simulation of the indoor parking lot. This simulation is intended to accurately simulate the parking area, including variations in  parking layouts, automobile kinds, and barriers such as pillars and walls. The video from these four viewpoints is then turned into photos, capturing both adequate and low-light circumstances in one shot. The dataset is created by annotating the required objects, specifically pillars and vehicles, from these photos. The dataset contains 150 photos, all of which have been scaled to 640x640 to ensure consistency. \\

\subsubsection{Processing Stage:}
The images were classified and labeled to train and test the object detection model. Several image labeling methods were investigated, and the Roboflow tool was finally chosen for object annotation. Roboflow creates a PASCAL VOC XML file with bounding box information, including xmax, xmin, ymax, and ymin coordinates. In the context of indoor parking infrastructure, recognizing automobiles and pillars is critical for obtaining information about available parking places. \\

After obtaining XML files containing information about the input images, all data was compiled into a CSV file, serving as the dataset for the object detection model. This CSV file was then split into two separate files: "$train_label.csv$" and "$test_label.csv$," allocating 75\% of the data for training and 25\% for testing purposes. The pre-trained YOLOv8 model was employed for object detection tasks, utilizing the annotated train and test files alongside their corresponding images as input. A model pipeline configuration file was generated, incorporating class information, batch size (set at eight), and the number of epochs (set at 50) for training the model. \\

After completing the training process, the model underwent testing using parking images to identify cars and pillars accurately. Fine-tuning of the model was performed, and an early stop method was implemented to halt training once the validation accuracy reached a plateau or showed signs of decline. This strategy was crucial in preventing overfitting and optimizing training efficiency by reducing unnecessary training iterations. \\

\begin{figure}[htbp]
    \centering
    \includegraphics[scale=1]{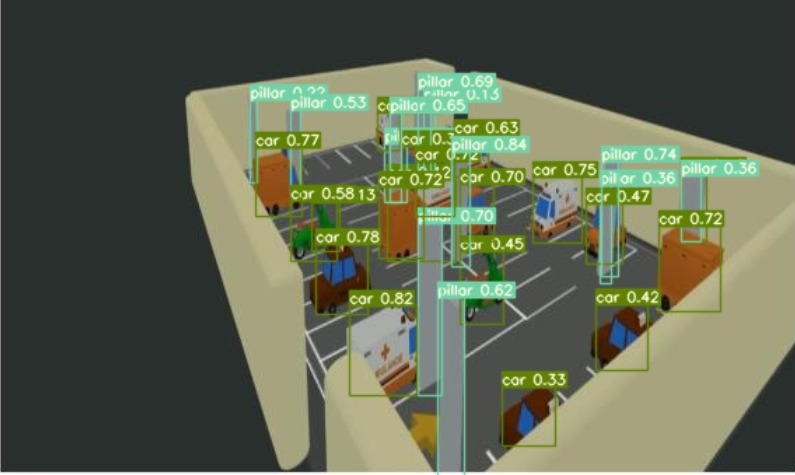}
    \caption{Vehicles and Pillar detection}
\end{figure}

The labelled images was converted into XML file to extract the bounding box coordinates for plotting. The bounding box coordinates are plotted in the XY plane and the depth is the Z coordinate.The following were the mathematical methods used to calculate the Z-coordinate.

\textbf{Centroid Euclidean Distance}
In our approach, we calculate the centroid of each object using the annotations CSV file, where the centroid coordinates are
\begin{gather}
    C(x_{centroid}, y_{centroid}) = (\frac{(x_{max} - x_{min})}{2} , \frac{(y_{max} - y_{min})}{2}) \label{one}
\end{gather}

This centroid computation is critical because it generates a representative point within the object's bounding box, which aids in spatial analysis. In addition, we use the camera as the origin, with coordinates (0,0), and calculate the Euclidean distance between the centroid and the camera. The distance calculation is represented by the Euclidean distance formula.

\begin{gather*}
    D_{xy} = \sqrt{(x_{centroid} - 0)^2 + (y_{centroid} - 0)^2} \label{two}
\end{gather*}

\textbf{Inversing Distance}
To get the z-coordinate representing the depth of an object within the parking lot, we use the inverse of the Euclidean distance (D) obtained from the camera to the centroid of the object. 

\begin{gather*}
    z = \frac{1}{D_{xy}} \label{three}
\end{gather*}

The reasoning behind this strategy is that when an item is closer to the camera, the inverse of the distance produces a greater value, resulting in a positive z-coordinate. Conversely, if the object is farther away, the inverse value will be smaller, resulting in a negative z coordinate. This logic is consistent with the idea that things closer to the observer have positive depth values, whilst those farther away have negative depths, allowing for spatial knowledge of the objects' placements inside the parking lot.

\subsubsection{Output Stage}
The 3d cartesian plotting process begins with analyzing collected data to determine which objects and their z-coordinates should be plotted in 3D Cartesian space. This involves computing the area of each annotated object's bounding box, favoring larger areas that are likely closer to other objects within the parking spot. Objects with smaller areas, indicating they are further away and less significant for the 3D plot from a particular camera position, are considered for scaling down or exclusion. This approach aids in combining photos from different camera perspectives effectively. \\

To handle overlapping sections in the 3D plot, the method calculates the area occupied by each object using bounding box coordinates from the annotated dataset. Determining the largest object class helps identify objects closest to the camera and ensures accurate depth computation. This information is then used to position objects in three dimensions, with smaller objects potentially scaled down or omitted to maintain clarity and accuracy. Utilizing matplotlib, the final 3D Cartesian plot integrates processed images from all camera angles, excluding overlapping regions for an accurate representation of pillars and vehicles in the parking space. \\

After plotting the pillars and vehicles in the 3D model, space extraction entails detecting and marking the residual spaces between two pillars or two vehicles as vacant parking spots. This technique is required to create an accurate picture of the parking lot and determine the availability of parking spaces. By evaluating the 3D plot and identifying vacant zones, our system can dynamically update and display real-time parking spot availability to users. \\

Jayaprakash et al. \cite{rp21} introduced a solution for 3D floor mapping of pillars and walls using a single camera view. In contrast, our proposed approach entails stitching together pictures from four cameras positioned at various angles to form a complete image. This method allows us to precisely assess parking spaces by measuring the distance between pillars along the x-axis and dividing it by the allotted parking space width, resulting in an estimate of the number of open spaces between two pillars. Furthermore, we address the problem of overlapped patches containing cars by eliminating them during the space extraction procedure. This approach improves the accuracy of locating vacant parking places while also optimizing parking space consumption.

\begin{figure}[htbp]
    \centering
    \includegraphics[scale=1.1]{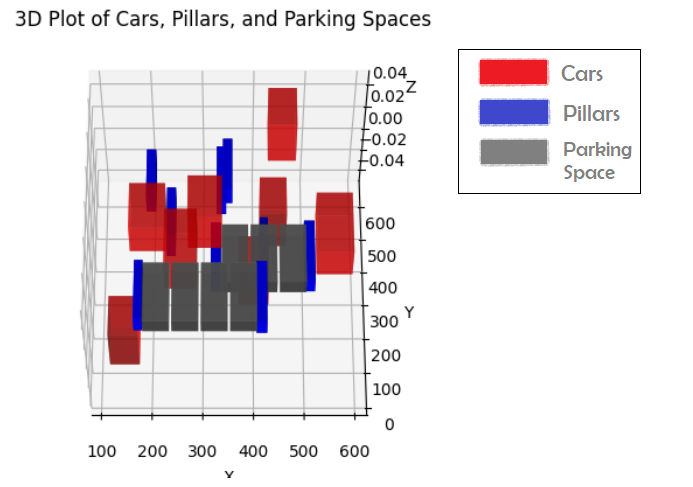}
    \caption{3D Cartesian plot indicating empty parked spaces}
\end{figure}

\section{Theoretical Analysis}
\textbf{Lemma 1} \textit{The Area-based method provides a more reliable and consistent depth estimation than the Centroid-based Inverse distance method.} \\
\textit{Proof: } Assuming the camera to be the origin with coordinates (0,0), let us consider $C(x_{centroid}, y_{centroid})$ to be the centroid of the object, $D_{xy}$ be the Euclidean distance between the camera and the centroid coordinates and $z$ be the z-coordinate by inversing the Euclidean distance $D_{xy}$. Now, let $A_{xy}$ be the area of the bounded box where 

\begin{gather}
    A_{xy} = ((x_{max} - x_{min}) * (y_{max} - y_{min})) \label{one}
\end{gather}
The corresponding depth $z'$ is calculated by taking the inverse of the area $A_{xy}$ where

\begin{gather*}
    z' = \frac{1}{A_{xy}} 
\end{gather*}

\textbf{Case 1:} Consider two objects at the same distance from the camera but with different areas. Let the centroids of the two objects be $C_1(x_{centroid1}, y_{centroid1})$ and $C_2(x_{centroid2}, y_{centroid2})$ and the corresponding Euclidean distances be $D_{x_{1}y_{1}}$ and $D_{x_{2}y_{2}}$ respectively. Since the objects are at the same distance from the camera,
\begin{gather*}
   D_{x_{1}y_{1}} = D_{x_{2}y_{2}} \\
   z_{1} = \frac{1}{D_{x_{1}y_{1}}} \\
   z_{2} = \frac{1}{D_{x_{2}y_{2}}} \\
   Hence, z_{1} = z_{2}
\end{gather*}

Now, using the area-based method, let the areas of 2 objects be $A_{x_{1}y_{1}}$ and  $A_{x_{2}y_{2}}$ respectively. Since the areas of the two objects are different, 

\begin{gather*}
   A_{x_{1}y_{1}} \neq A_{x_{2}y_{2}} \\
   z'_{1} = \frac{1}{A_{x_{1}y_{1}}} \\
   z'_{2} = \frac{1}{A_{x_{2}y_{2}}} \\
   Hence, z'_{1} \neq z'_{2}
\end{gather*}
This contradicts the assumption that the Area-based method provides more consistent depth values for objects. 

\textbf{Case 2:} Consider an object moving directly away from the camera while maintaining its size. Here the Centroid-based method shows a decrease in depth $z$ when the object moves farther whereas the Area-based method will not change the depth as the area of the object is consistent, so $z'$ will be the same. This contradicts the assumption that the Area-based method is more sensitive to the actual distance.

Hence in both the cases the Centroid-based method proves to be superior in reliable depth estimation, thereby contradicting the assumption.

\section{Results and Analysis}

Object detection tasks frequently contain imbalanced data, with much fewer positive examples (items of interest) than negative occurrences. In such instances, depending exclusively on accuracy numbers may be misleading. The PR curve addresses this by taking into account both accuracy and recall, providing a more nuanced picture of the model's ability to accurately identify positive events in the face of imbalanced data.

\begin{itemize}
    \item \textbf{Precision:} identifying positives among predicted positives
    \item \textbf{Recall:} identifying positives among actual positives
\end{itemize}

Here is a comparison of YOLOv5, YOLOv7, and YOLOv8's precision-recall curves. 

\begin{figure}[htbp]
    \centering
    \includegraphics[scale=0.4]{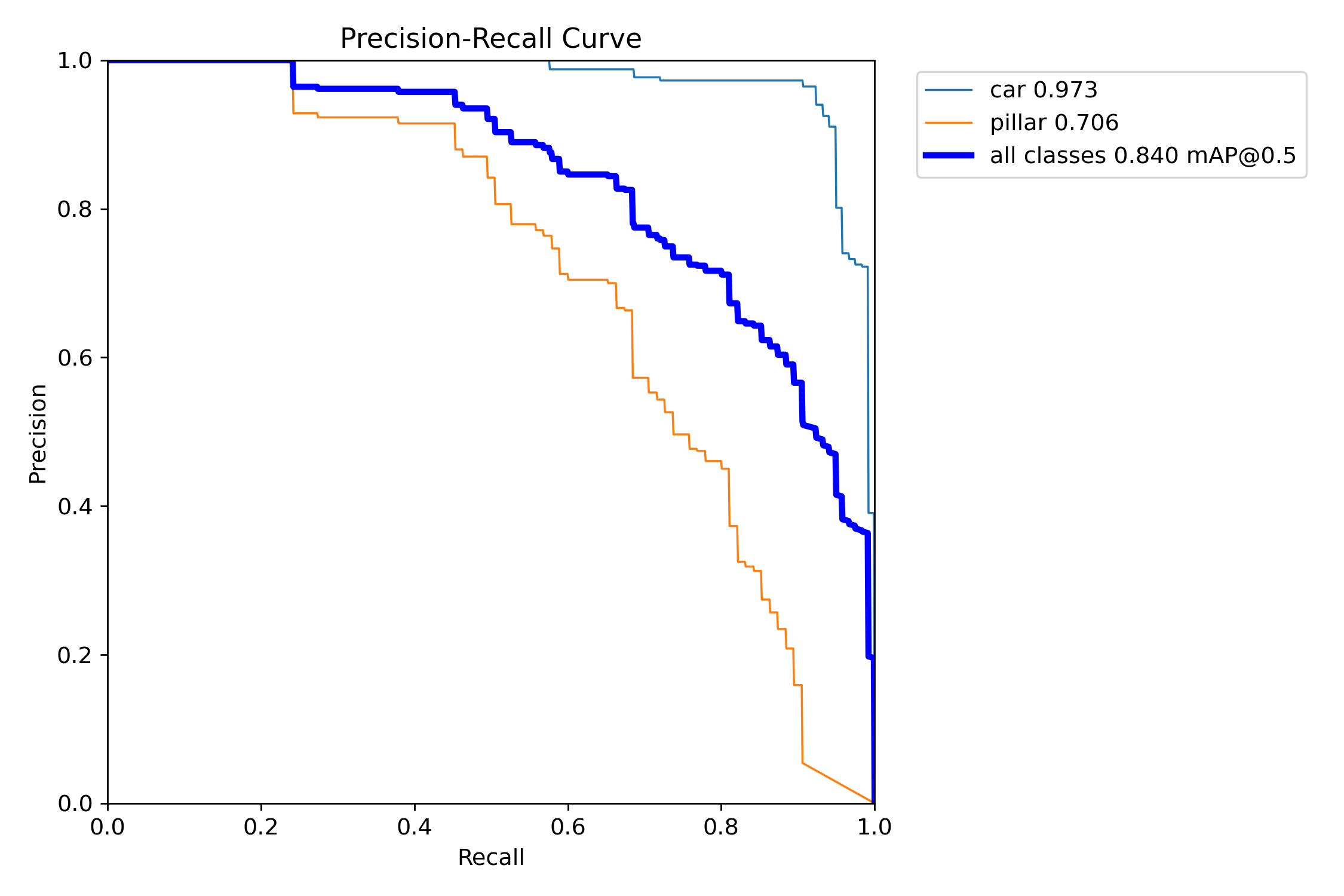}
    \caption{Precision-Recall Curve of YOLOv5 model}
\end{figure}

The precision-recall curve in the image visualizes the trade-off between accurately catching true positives (recall) and minimizing false positives (precision) in your classification model. The ideal scenario is in the upper right corner (high precision, high recall) but uncommon. A typical curve shows a trade-off, where increasing recall inevitably leads to some false positives. In Fig 7 we can see that YOLOv5, while still good, has an accuracy of 84\%, a car detection accuracy of 97.3\%, and a substantially lower pillar identification accuracy of 70.6\%, for the same number of epochs.

\begin{figure}[htbp]
    \centering
    \includegraphics[scale=0.4]{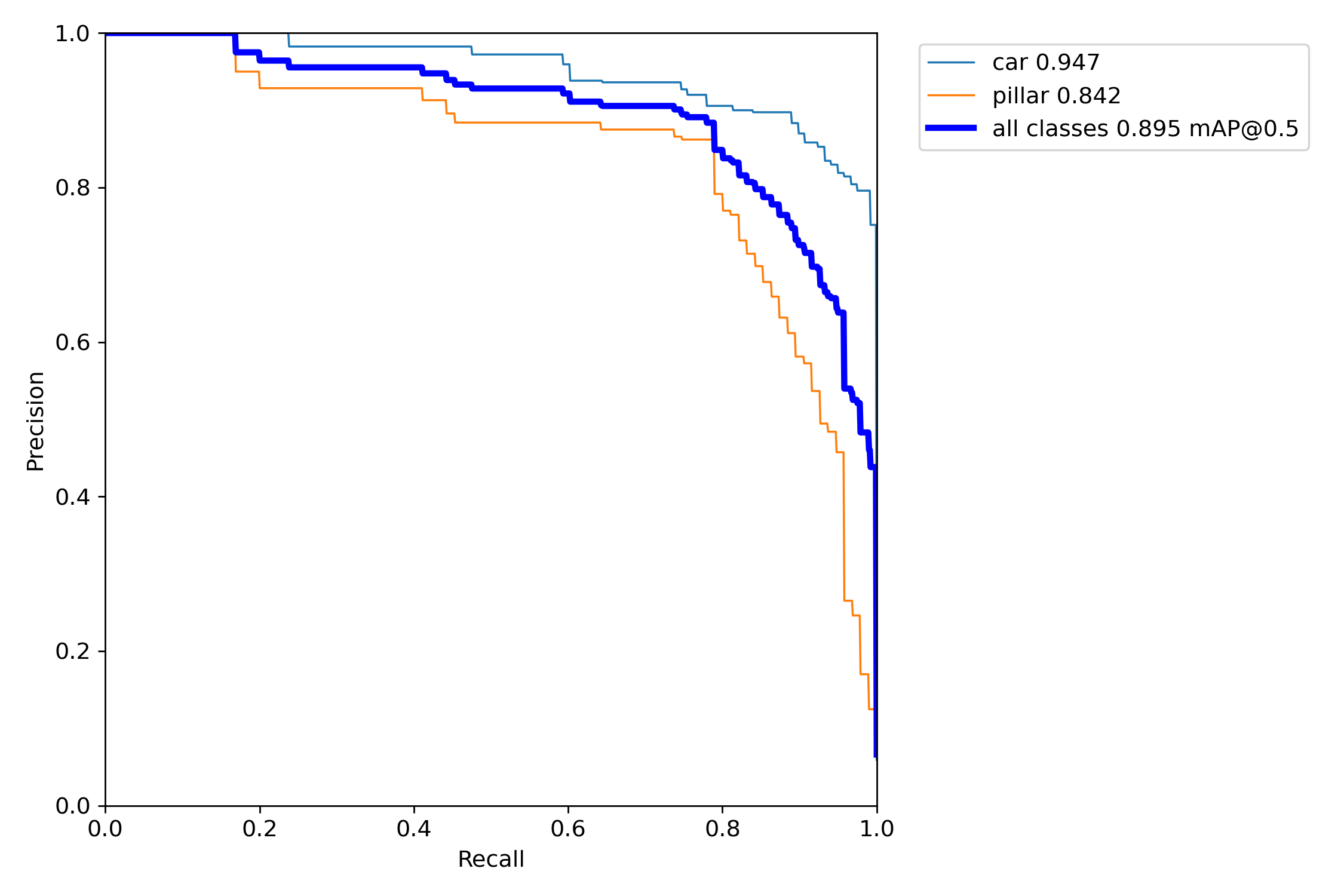}
    \caption{Precision-Recall Curve of YOLOv7 model}
\end{figure}

A steeper drop in precision indicates that YOLOv5 is less efficient in balancing the trade-off between recall and precision compared to YOLOv7. As illustrated in Fig 8, YOLOv7 demonstrates superior performance, achieving an overall accuracy of 89.5\%. Notably, it excels in car detection with an impressive accuracy of 94.7\%. However, its performance in pillar detection is slightly better than YOLOv5, with an accuracy of 84.2\%. Despite this discrepancy, the higher precision of YOLOv7 means that it is better at accurately distinguishing between cars and pillars, reducing the likelihood of false positives.

\begin{figure}[htbp]
    \centering
    \includegraphics[scale=0.4]{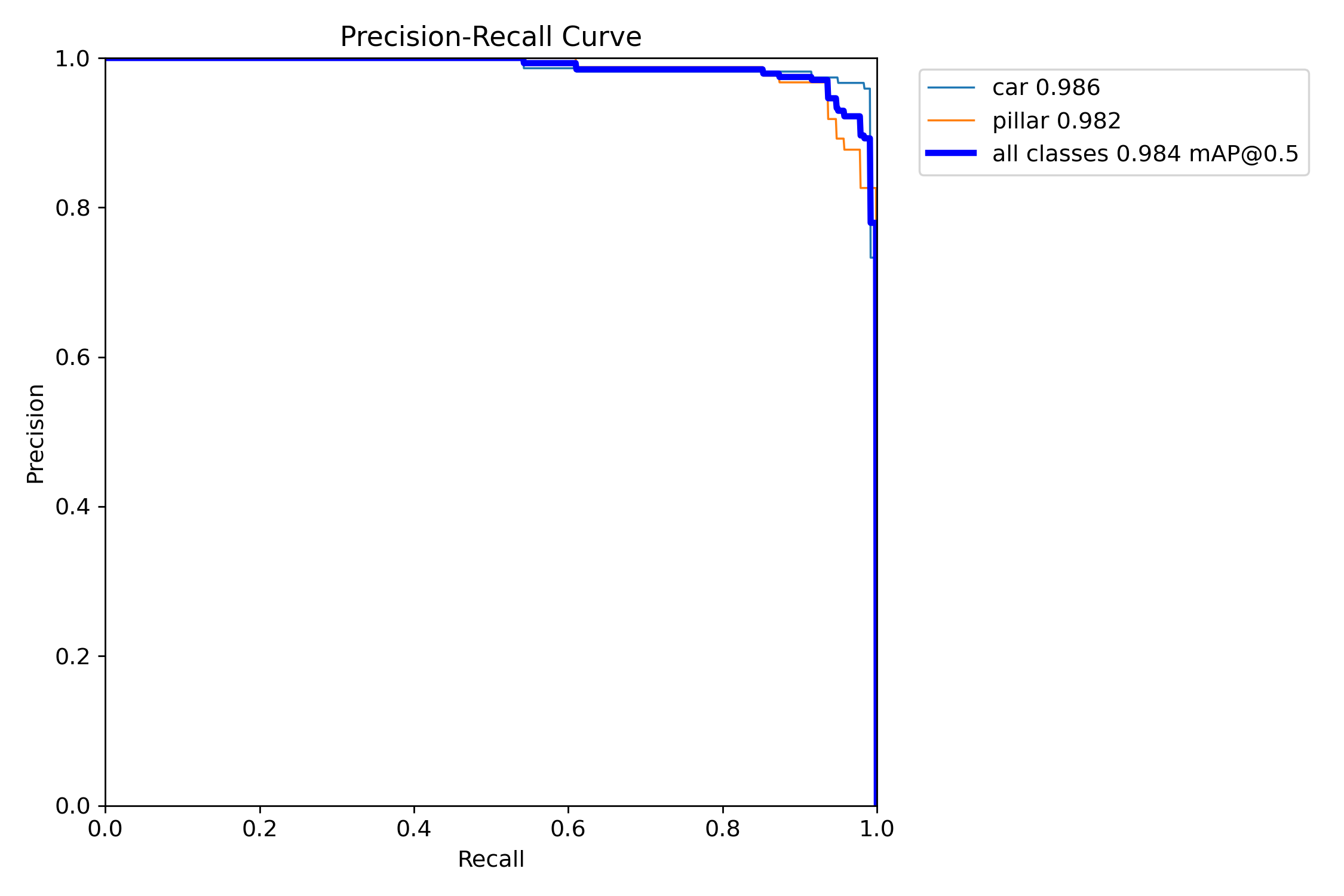}
    \caption{Precision-Recall Curve of YOLOv8 model}
\end{figure}

In Fig 9, YOLOv8 surpasses both YOLOv5 and YOLOv7, achieving the highest overall accuracy of 98.4\%. It maintains an excellent balance in detection performance, with 98.6\% accuracy for car detection and 98.2\% for pillar detection. This high level of accuracy across both categories underscores YOLOv8's robustness and precision, making it the most reliable model for accurate object classification tasks.

\section{Conclusion and Future Works}
In conclusion, the research in the field of smart cities underscores the critical role of data-driven solutions and sustainable infrastructure in shaping the urban landscape. Intelligent parking systems are crucial to encouraging sustainable transportation and easing urban congestion in the context of a smart city. However, challenges involving sensor precision and complex integration offer key obstacles to the efficient application of these systems. To address these challenges, our proposed work introduces an affordable and user-friendly computer vision-based smart parking system, leveraging advanced object detection models like YOLOv8. We can create combined image of parking lot from 4 camera placements using the novel inverse perspective mapping approach (IPM), which makes it possible to extract information about empty parking spaces quickly and effectively using a 3D Cartesian plot. \\

Future research should concentrate on enhancing and perfecting the suggested system for use in practical applications. This includes tackling potential scalability problems in larger indoor parking lots as well as further improving the computer vision algorithms' accuracy and reliability. The system's functionality and user experience can also be improved by taking into account IoT integration, data analytics, and user-friendly mobile applications. We can help create smarter, more sustainable urban settings by constantly improving smart parking technologies.

\bibliography{references}

@article{rp1,
  title={Smart parking system (SPS) architecture using ultrasonic detector},
  author={Kianpisheh, Amin and Mustaffa, Norlia and Limtrairut, Pakapan and Keikhosrokiani, Pantea},
  journal={International Journal of Software Engineering and Its Applications},
  volume={6},
  number={3},
  pages={55--58},
  year={2012},
  publisher={Science and Engineering Research Support Society Australia}
}

@inproceedings{rp7,
  title={Automatic parking space detection system},
  author={Bibi, Nazia and Majid, Muhammad Nadeem and Dawood, Hassan and Guo, Ping},
  booktitle={2017 2nd international conference on multimedia and image processing (ICMIP)},
  pages={11--15},
  year={2017},
  organization={IEEE}
}

@article{rp15,
  title={T-YOLO: Tiny vehicle detection based on YOLO and multi-scale convolutional neural networks},
  author={Carrasco, Daniel Padilla and Rashwan, Hatem A and Garc{\'\i}a, Miguel {\'A}ngel and Puig, Domenec},
  journal={IEEE Access},
  year={2021},
  publisher={IEEE}
}

@article{rp17,
  title={Computer Vision Based Parking Optimization System},
  author={Chandrasekaran, Siddharth and Reginald, Jeffrey Matthew and Wang, Wei and Zhu, Ting},
  journal={arXiv preprint arXiv:2201.00095},
  year={2022}
}

@article{rp20,
  title={Camera-Based Smart Parking System Using Perspective Transformation},
  author={Liu, Bowie and Lai, Hawking and Kan, Stanley and Chan, Calana},
  journal={Smart Cities},
  volume={6},
  number={2},
  pages={1167--1184},
  year={2023},
  publisher={MDPI}
}

@inproceedings{rp21,
  title={Computer Vision Based 3D Model Floor Construction for Smart Parking System},
  author={Patra, Jayaprakash and Panda, Satyajit and Negi, Vipul Singh and Chinara, Suchismita},
  booktitle={IFIP International Internet of Things Conference},
  pages={36--48},
  year={2023},
  organization={Springer}
}

@INPROCEEDINGS{rp22,
  author={Qiang, MAI and Jiaming, LIANG and Qinfeng, ZHANG and Chunliang, HSU},
  booktitle={2019 IEEE International Conference on Computation, Communication and Engineering (ICCCE)}, 
  title={Indoor Parking Navigation System Using Visible LED Light Communication}, 
  year={2019},
  volume={},
  number={},
  pages={157-159},
  doi={10.1109/ICCCE48422.2019.9010803}}

@INPROCEEDINGS{rp23,
  author={Bojja, J. and Kirkko-Jaakkola, M. and Collin, J. and Takala, J.},
  booktitle={2013 IEEE International Conference on Acoustics, Speech and Signal Processing}, 
  title={Indoor 3D navigation and positioning of vehicles in multi-storey parking garages}, 
  year={2013},
  volume={},
  number={},
  pages={2548-2552},
  doi={10.1109/ICASSP.2013.6638115}}

@ARTICLE{rp24,
  author={Gong, Zheng and Li, Jonathan and Luo, Zhipeng and Wen, Chenglu and Wang, Cheng and Zelek, John},
  journal={IEEE Transactions on Intelligent Transportation Systems}, 
  title={Mapping and Semantic Modeling of Underground Parking Lots Using a Backpack LiDAR System}, 
  year={2021},
  volume={22},
  number={2},
  pages={734-746},
  doi={10.1109/TITS.2019.2955734}}

@INPROCEEDINGS{rp25,
  author={Du, Yichen and Arslan, Tughrul and Juri, Arief},
  booktitle={2016 International Conference on Indoor Positioning and Indoor Navigation (IPIN)}, 
  title={Camera-aided region-based magnetic field indoor positioning}, 
  year={2016},
  volume={},
  number={},
  pages={1-7},
  doi={10.1109/IPIN.2016.7743621}}

@ARTICLE{rp26,
  author={Cai, Bill Yang and Alvarez, Ricardo and Sit, Michelle and Duarte, Fábio and Ratti, Carlo},
  journal={IEEE Internet of Things Journal}, 
  title={Deep Learning-Based Video System for Accurate and Real-Time Parking Measurement}, 
  year={2019},
  volume={6},
  number={5},
  pages={7693-7701},
  doi={10.1109/JIOT.2019.2902887}}

@inproceedings{rp27,
  title={An integration system of AI cars detection with enclosed photogrammetry for indoor parking lot},
  author={Li, Haoxuan and Wu, Weihong and Li, Yingze},
  booktitle={The 2nd International Conference on Computing and Data Science},
  pages={1--6},
  year={2021}
}

@article{rp28,
  title={Intelligent parking guidance system using computer vision, IoT, and edge computing},
  author={Rajesh, Pranav and Kallakuri, Dheeraj and Chagarlamudi, Sudeeksha and Vissavajhula, Sai Manoj and Meng, Jingxiong and Zhao, Junfeng},
  journal={IEEE Open Journal of Intelligent Transportation Systems},
  year={2025},
  publisher={IEEE}
}

@inproceedings{rp29,
  title={Dijkstra's algorithm and Google maps},
  author={Lanning, Daniel R and Harrell, Gregory K and Wang, Jin},
  booktitle={Proceedings of the 2014 ACM Southeast Conference},
  pages={1--3},
  year={2014}
}

@inproceedings{rp30,
  title={A necessary review on optical character recognition (OCR) system for vehicular applications},
  author={Bansal, Shivani and Gupta, Meenu and Tyagi, Amit Kumar},
  booktitle={2020 Second International Conference on Inventive Research in Computing Applications (ICIRCA)},
  pages={918--922},
  year={2020},
  organization={IEEE}
}

@incollection{rp31,
  title={Advanced smart parking management system integrating deep learning and optical character recognition},
  author={Bhaskar, Navaneeth and Waghmare, Priyanka Tupe and Puttur, Ashritha Kalluraya and Pereira, Ovin Vinol},
  booktitle={Advances in Electronics, Computer, Physical and Chemical Sciences},
  pages={355--360},
  year={2025},
  publisher={CRC Press}
}

@article{rp32, 
title={Yolo versions architecture}, 
author={Hasan, Rusul Hussein and Hassoo, Rasha Majid and Aboud, Inaam Salman}, 
journal={International Journal of Advances in Scientific Research and Engineering}, 
volume={9}, 
number={11}, 
pages={73}, 
year={2023} 
}

@article{rp33,
  title={Challenges and opportunities in smart parking sensor technologies},
  author={Shroud, Mohamed Abu and Eame, Masbah and Elsaghayer, Eltuhami and Almabrouk, Abdulsalam and Nassar, Yasser},
  journal={Int. J. Electr. Eng. and Sustain.},
  pages={44--59},
  year={2023}
}

@article{rp34,
  title={An overview of autonomous parking systems: Strategies, challenges, and future directions},
  author={Olmos Medina, Javier Santiago and Maradey L{\'a}zaro, Jessica Gissella and Rass{\~o}lkin, Anton and Gonz{\'a}lez Acu{\~n}a, Hern{\'a}n},
  journal={Sensors},
  volume={25},
  number={14},
  pages={4328},
  year={2025},
  publisher={MDPI}
}

@inproceedings{rp35,
  title={Challenges in visual parking and how a developmental network approaches the problem},
  author={Zheng, Zejia and Weng, Juyang},
  booktitle={2016 International Joint Conference on Neural Networks (IJCNN)},
  pages={4593--4600},
  year={2016},
  organization={IEEE}
}

@article{rp36,
  title={A review of smart parking systems},
  author={Channamallu, Sai Sneha and Kermanshachi, Sharareh and Rosenberger, Jay Michael and Pamidimukkala, Apurva},
  journal={Transportation Research Procedia},
  volume={73},
  pages={289--296},
  year={2023},
  publisher={Elsevier}
}

\bibliographystyle{unsrt}  

\end{document}